\documentclass[letterpaper, 10 pt, conference]{ieeeconf}  %

\IEEEoverridecommandlockouts                              

\usepackage{graphics} 
\usepackage{epsfig} 
\usepackage{mathptmx} 
\usepackage{times} 
\usepackage{amsmath} 
\usepackage{amssymb}  

\usepackage{amsthm}
\usepackage[T1]{fontenc}
\usepackage{subcaption}
\usepackage{color}
\usepackage{url}

\newtheorem{thm}{Theorem}[section]
\newtheorem{lem}[thm]{Lemma}

\title{\LARGE \bf
Constrained Dynamic Movement Primitives for \\Safe Learning of Motor Skills
}

\author{Seiji Shaw$^\dagger$, Devesh K. Jha$^\ddagger$, Arvind U. Raghunathan$^\ddagger$, Radu Corcodel$^\ddagger$, \\Diego Romeres$^\ddagger$, George Konidaris$^\dagger$ and Daniel Nikovski$^\ddagger$
\thanks{$^\dagger$Department of Computer Science, Brown University, Providence, Rhode Island. {\tt\small \{sshaw4,gdk\}@cs.brown.edu}. $^\ddagger$ Mitsubishi Electric Research Labs (MERL), Cambridge, MA USA.
{\tt\small lastname@merl.com}}%
}

\newcommand{\Int}[0]{\text{Int}}

\begin{document}

\maketitle
\thispagestyle{empty}
\pagestyle{empty}

\begin{abstract}
Dynamic movement primitives are widely used for learning skills which can be demonstrated to a robot by a skilled human or controller. While their generalization capabilities and simple formulation make them very appealing to use, they possess no strong guarantees to satisfy operational safety constraints for a task. In this paper, we present constrained dynamic movement primitives (CDMP) which can allow for constraint satisfaction in the robot workspace. We present a formulation of a non-linear optimization to perturb the DMP forcing weights regressed by locally-weighted regression to admit a Zeroing Barrier Function (ZBF), which certifies workspace constraint satisfaction. We demonstrate the proposed CDMP under different constraints on the end-effector movement such as obstacle avoidance and workspace constraints on a physical robot. A video showing the implementation of the proposed algorithm using different manipulators in different environments could be found here \url{https://youtu.be/hJegJJkJfys}. 
\end{abstract}

\section{Introduction}

Programming robotic arm manipulators for tasks that require complex, dynamic movements is often difficult and time-consuming. Learning from Demonstration (LfD) offers a more efficient approach to teach motor skills to robots~\cite{argall2009survey, ravichandar2020recent}, since expert demonstrations can be directly used to learn a suitable representation of the movement to solve the given task. However, it is desirable that these approaches can provide some guarantees regarding operational constraints, which are more difficult to specify from expert demonstration. 

One of the most popular LfD approaches is the dynamic movement primitive (DMP), a nonlinear dynamical system that can fit expert demonstration trajectories by decoupling a nonlinear forcing function from the nominal attraction behavior. Since their introduction by Schaal et al, \cite{schaal2006dynamic}, DMPs have been applied to learning a wide range of skills \cite{saveriano2021dynamic, 5686298, mulling2013learning, 9838102, sharma2019learning}. While they can be reparameterized by their start and goal positions and preserve the qualitative shape of the trajectory, incorporating arbitrary operational constraints on the trajectory is very difficult. Without care, a new start or goal pose may cause the robot to collide with itself or the environment. Consequently, working with DMPs in arbitrary environments require careful engineering to ensure that the computed trajectories stay collision-free and within workspace bounds. 

For dynamical systems like DMPs, safety can be defined by whether the trajectories of the system remain within a specified `safety set.' Such a set can be defined by removing obstacles from the set that defines the robot's workspace or by joint inequality constraints derived joint limits on a manipulator arm. Ames et al. \cite{ames2016control} introduce Zeroing-Barrier Functions (ZBFs), a formalism that can certify whether a dynamical system is \textit{forward-invariant} in a safety set, i.e. whether the state of the system is always in the safety set. A key challenge is to find a suitable barrier function that can be used to certify satisfaction of all constraints desired by the user. 

\begin{figure*}
\centering
\includegraphics[scale=0.75]{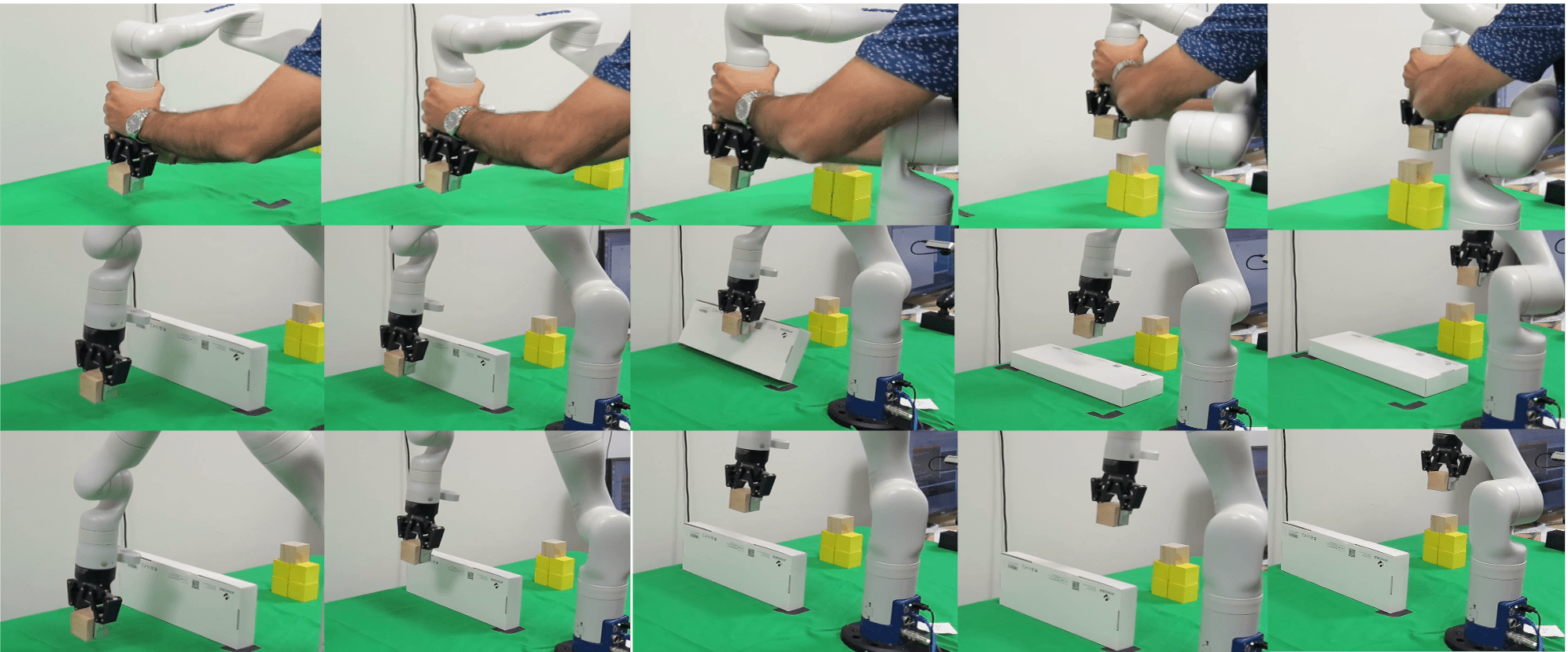}
\caption{A simple block stacking environment using the proposed CDMP algorithm. The demonstration (top row) was provided in a collision-free environment. A box obstacle is introduced during execution and a simple DMP collides (middle row) with the obstacle during execution. The proposed CDMP (bottom row) is able to successfully avoid the novel obstacle.}
\label{fig:cdmp_obstacle_avoidance}
\end{figure*}

We present Constrained DMPs (CDMPs) which allow constraint satisfaction for the motor skills learned using DMPs.
Given a user-defined ZBF that specifies the safety set in the robot workspace, we optimize a set of parameters to the DMP's nonlinear forcing function to guarantee that the DMP trajectory lies within the safety set. We cast the optimization as a nonlinear program which can be solved using off-the-shelf non-linear program solvers such as IPOPT~\cite{wachter2006implementation, DBLP:journals/corr/abs-2106-03220}. Figure~\ref{fig:cdmp_obstacle_avoidance} shows a motivating example where an initial demonstration was provided in an obstacle-free environment. The DMP computed using the demonstration collides with a novel obstacle during execution. The algorithm proposed in this paper is able to compute a collision-free trajectory. 

\textbf{Contributions.} This paper presents the following contributions:
\begin{enumerate}
    \item A novel formulation for Constrained DMPs which can satisfy operational constraints to ensure safe reproduction of motor skills using demonstrations.
    \item Demonstrations on several examples where we can successfully compute CDMPs which satisfy operational constraints in the robot workspace.
\end{enumerate}
Compared to the past work on incorporating collision avoidance constraints in DMPs ~\cite{rai2014learning, park2008movement, tan2011potential, stulp2009compact}, we are able to provide guarantees for safety set invariance for the new trajectories, which encompass obstacle-avoidance, workspace constraints, etc. 


\section{Related Work}\label{sec:related_work}



Learning from demonstration, or imitation learning, is an active area of research in the robot learning community~\cite{argall2009survey}. In general, there are two common behavior representations that are trained by LfD:  a stochastic policy or a dynamical or geometric representation of a concrete trajectory. A case of the latter, DMPs have been widely successful in learning motor skills for robots because they present a simple dynamical system that can encode a wide range of motor skills.

DMPs enjoy wide application in a variety of robot tasks, but are unable to incorporate operational constraints with strong guarantees. Collision avoidance is usually incorporated by combining DMPs with artificial potential fields (or using other similar heuristics)~\cite{rai2014learning, park2008movement}. In general, the technique lacks a formulation that can naturally incorporate any form of positional constraint without the time-consuming experimental tuning processes or risks of local minima that artificial potential field-like approaches possess (\cite{park2008movement}, Sec. \ref{sec:apf}). More recently, there has been some work on using DMPs with sampling based methods to allow collision avoidance~\cite{sobti2021sampling}. These methods use sampling to find a state from which the DMP can avoid obstacles. In contrast, in the proposed method, we perturb the learned forcing function to satisfy constraints on the robot's workspace.

We also draw inspiration from approaches in safety-critical control that uses optimization-based approaches to constrain the control inputs of a dynamical system to maintain strong guarantees of the system's safety. At the center of these approaches are control Lyapunov functions (CLFs) \cite{ames2014rapidly} and control barrier functions (CBFs) \cite{ames2016control, ames2019control, romdlony2014uniting}, which certify convergence and safety set forward-invariance, respectively. In a similar vein, Tedrake et al. \cite{tedrake2010lqr} compute Lyapunov-stable regions for locally-defined LQR controllers via sum-of-squares optimization. Since a DMP is just a nonlinear dynamical system, we leverage Ames et al.'s zeroing barrier function (ZBF) formalism (which Ames et al.  extends to control-affine systems to define CBFs) to certify the forward-invariance of a DMP in a predetermined safety set \cite{ames2016control, ames2019control}. 

A key insight of our work is to perturb the weights of an already-learned DMP to admit an already-existing ZBF so that the DMP is guaranteed to be forward-invariant in a user-defined safety set. There is a large body of work that constrains more general parameterizable dynamical systems via Lyapunov functions \cite{khansari2011learning, khansari2014learning, figeuroa2017physically} and contraction mappings \cite{ravichandar2017learning}. While these techniques guarantee convergence of the dynamical system, such methods cannot maintain the DMP's ability to generalize motions over changes in the attraction point, a property that our approach preserves. Nomotista et al. \cite{notomista2021safety} leverage a diffeomorphism to map non-convex obstacles into a complex space, where they constrain dynamics of a control system via CBF. Khansari-Zadeh et al. \cite{khansari2012dynamical} propose a  real-time dynamical-reshaping approach for avoidance of convex obstacles. While these approaches work for obstacles that follow their assumptions, it is unclear if such an approach can be modified for avoidance of non-compact sets as well. Ohnishi et al. \cite{ohnishi2021constraint} consider CBFs for learning agents to avoid more abstract task-specific constraints for more agents whose dynamics are learned via reinforcement learning. We solve a task-specific learning-by-demonstration problem instead.


\section{Background}
In this section, we review the mathematical details of DMPs and ZBFs relevant to our construction of CDMP.

\subsection{Dynamic Movement Primitives}\label{sec:DMP}

DMPs were first introduced by Schaal et al.~\cite{schaal2006dynamic}. To remove explicit time dependency, they use a canonical system to keep track of the progress through the learned behavior:
\begin{equation}
\tau \dot s = -\alpha_s s
\end{equation}

where $s = 1$ at the start of DMP execution (and $\alpha_s > 0$) and $\tau > 0$ specifies the rate of progress through the DMP.

To capture attraction behavior, DMPs use a spring-damper system (the transformation system) with an added nonlinear forcing term. Writing the DMP equations as a system of coupled first-order ODEs yields:

\begin{align}
    \tau \dot z &= \alpha_z(\beta_z (g - y) - z) + f(s)\\ 
    \tau \dot y &= z 
\end{align}

where $g$ denotes the goal pose. The forcing term is defined as a radial-basis function:

\begin{align}
    f(s) &= \frac{\sum_{i=1}^N w_i \psi_i(s)}{\sum_{i=1}^N \psi_i(s)}\\
    \psi_i(s) &= \exp{(-h_i (s-c_i)^2)}
\end{align}

where $h_i$  and $c_i$ denote the width and center of the Gaussian basis functions, respectively. The forcing term is learned from the demonstration by solving a locally weighted regression to fit the demonstration provided by an expert. 

\subsection{Zeroing Barrier Functions}

Since DMPs are formulated as an autonomous nonlinear dynamical system, we can certify them for safety set forward-invariance using Ames et al.'s zeroing barrier function formalism \cite{ames2019control, ames2016control}. 


First, let $h(x): \mathbb R^n \to \mathbb R$ be a continuous differentiable function. Define set $\mathcal C$ to be the following super-level set:

\begin{align}
    \mathcal C &= \{x \in \mathbb R^n: h(x) \geq 0\}\\
    \partial \mathcal C &= \{x \in \mathbb R^n : h(x) = 0\}\\
    \Int(\mathcal C) &= \{x \in \mathbb R^n : h(x) > 0\}
\end{align}


Assume we have a nonlinear dynamical system of the form:
\begin{equation}
    \label{eq:affine_ds}
    \dot x = f(x)
\end{equation}




\noindent We call $h(x)$ ZBF if the following inequality holds:

\begin{equation}
    \dot h(x) \geq -\alpha(h(x))
\end{equation}

\noindent where $\alpha: \mathbb R \to \mathbb R$ is a class-$\mathcal K$ function \cite{ames2016control}. Like Ames et al., we consider the specific case 
\begin{equation}
    \dot h(x) \geq -\gamma h(x).
\end{equation}

\noindent with constant $\gamma > 0$. Given a dynamical system (\ref{eq:affine_ds}), if we can find a valid ZBF $h(x)$(or alternatively, change our dynamical system to admit an existing ZBF), then our dynamical system $\dot x = f(x)$ is forward-invariant in the set $\mathcal C$, i.e. if $x_0 \in \mathcal C$ and $\dot x = f(x)$ then:
\begin{equation}
    x \in \mathcal C, \mbox{ } \forall t \in [0, \infty)
\end{equation}

\section{Constrained Dynamic Movement Primitives}
We now present our formulation of the constrained DMP, which we cast as a nonlinear optimization problem.


The primary insight behind our proposed formulation of CDMPs is to take an existing DMP with a forcing function learned from an expert trajectory, and then optimize perturbations to this forcing term (Sec. \ref{subsec:cdmp_statement}) so that the DMP dynamical system admits a ZBF that certifies trajectory generation within a pre-defined safety set. This safety set (and ZBF) is constructed by composing signed-distance fields from primitive convex polytopes (Sec. \ref{subsec:zbs_from_sdfs}). 

\subsection{CDMPs as a Nonlinear Optimization Problem}\label{subsec:cdmp_statement}

As was explained earlier in Section~\ref{sec:DMP}, the system of equations and the forcing function for DMP could be written and expressed by the following system of equations:


\begin{equation}
\label{eqn:tensor_dmp}
\begin{bmatrix}
   \dot s \\
   \dot z \\
   \dot y \\
\end{bmatrix}
=
\frac{1}{\tau}
\begin{bmatrix}
    -\alpha_s s\\
    \left(\alpha_z(\beta_z (g - y) - z) +f(s)\right)\\
    z \\ 
\end{bmatrix}
\end{equation}
where 
\begin{equation}
    f(s) = \frac{\sum_{i=1}^N w_i \psi_i(s)}{\sum_{i=1}^N \psi_i(s)}
\end{equation}
is the forcing function expressed as sum of radial basis functions and which are learned from the provided expert demonstration. 



We introduce additional parameters to the DMP dynamical systems that can be optimized to allow constraint satisfaction. In the proposed formulation, CDMPs are computed by optimizing a set of perturbations $\{\zeta_i\}$ of the original (regressed) weights $\{w_i\}$ of the learning forcing function to obey the ZBF inequality for forward-invariance in the user-defined safety set. More formally, we define the perturbed forcing function as follows:


\begin{equation}
    \tilde{f}(s) = \frac{\sum_{i=1}^N (w_i - \zeta_i) \psi_i(s)}{\sum_{i=1}^N \psi_i(s)}
\end{equation}
where $\zeta_i$ are the decision variables optimized for the formulated optimization problem. We note that this is only one possible way to perturb the DMP forcing function, which we based on its most common representation as a radial-basis function in the DMP literature. 


We now introduce the formulation of the nonlinear optimization problem to compute the parameter set $p = \{\zeta_i\}$:



\begin{equation}\label{eqn:cdmp}
    \min_{x, p} \int_{t_0}^{t_f} c(x(t), p)dt + \phi(x(t_f), p)
\end{equation}

\noindent subject to the dynamic constraints


\begin{align}
\tau
\begin{bmatrix}
   \dot s \\
   \dot z \\
   \dot y \\
\end{bmatrix}
+
\begin{bmatrix}
    \alpha_s s\\
    -\left(\alpha_z(\beta_z (g - y) - z) + \tilde{f}(s)\right)\\
    -z \\ 
\end{bmatrix}
&=
\mathbf{0}\label{eqn:cdmp_dynamics}\\ 
\dot h(x) + \gamma h(x) &\geq 0 \label{eqn:zbf_constraint} \\ 
x(t_0) = x_0 \label{eqn:init_condition}
\end{align}

\noindent where $x = [s, z, y]^T$ and $p$ is the set of parameters (including the $\zeta_i$). Note that the function $c(\cdot,\cdot,\cdot)$ represents the cost functional for the optimization problem. The simplest cost function can penalize the L2-norm of the decision variables $\zeta_i$.

The problem formulation represented by Equations~\eqref{eqn:cdmp}-\eqref{eqn:init_condition} is converted to a finite dimensional discretized problem using PyRoboCOP~\cite{DBLP:journals/corr/abs-2106-03220}, which is then transcribed into a nonlinear program. The resulting NLP is then solved using IPOPT in the PyRoboCOP environment.

\subsection{Perturbation from Original Trajectory}

Another important consideration during computation of constrained DMP is to limit the amount of the deviation of the CDMP from the original DMP trajectory. The design of CDMP could be seen as a trade-off between constraint satisfaction and the original forcing function. This trade-off could be controlled using a hyperparameter which constrains the maximum allowable deviation between the original DMP and CDMP, which can be added as another constraint to the trajectory optimization problem. More formally, let us denote the original DMP trajectory as $\{\tilde{y}(t)\}$, $t\in [t_0,t_f]$ and the hyperparameter for the deviation from the original trajectory as $\epsilon$. Then, the additional constraint could be represented as follows:
\begin{equation}\label{eqn:traj_deviation}
    ||y(t)-\tilde{y}(t)||_2\leq \epsilon
\end{equation}
With this additional deviation constraint, the optimization problem can then be written as follows:
\begin{equation}
    \min_{x, p} \int_{t_0}^{t_f} c(x(t), p)dt + \phi(x(t_f), p)
\end{equation}
s.t.,       \eqref{eqn:cdmp_dynamics}-\eqref{eqn:init_condition} \\
\begin{equation}\label{eqn:traj_deviation_constraint}
    ||y(t)-\tilde{y}(t)||_2\leq \epsilon
\end{equation}
Depending on the value of $\epsilon$, we can obtain a family of CDMPs that will allow different amount of deviation of the new trajectory from the original DMP trajectory.

\subsection{ZBFs from Signed-Distance Functions} \label{subsec:zbs_from_sdfs}

Given an obstacle set $\Omega$ with boundary $\partial \Omega$ in $\mathbb R^3$, we use the usual definition of a signed distance function (SDFs) $\sigma: \mathbb R^3 \to \mathbb R$ to the boundary of the obstacle \cite{jones20063d}:

$$
\sigma(x) = \begin{cases}
d(x, \partial \Omega), & \text{if } x \notin \Omega,\\
-d(x, \partial \Omega), & \text{if } x \in \Omega.
\end{cases}
$$

\noindent and where the distance (or metric) $d(x, \partial \Omega)$ is defined by

$$
d(x, \partial \Omega) = \inf_{y \in \partial \Omega} d(x, y).
$$

\newcommand{\smin}{\operatorname{smin}}
\noindent We know that $|\nabla \sigma| = 1$ wherever this gradient is well-defined (and for convex polytopes, this is true whenever $\sigma(x) > 0$). We also know that the implicit surface embedded in $\mathbb R^3$ defined by the set of all points $x \in \mathbb R^3$ where $\sigma(x) = 0$ is also the boundary of the obstacle, and $\sigma(x) < 0$ whenever $x$ is inside the obstacle. Thus, SDFs fulfill the requirements for a function to designate a safety set (what we denote as $h(x)$ above) and certify that a DMP trajectory is also invariant in the safety set $\Omega^c$.
Given two SDFs $\sigma_1$ and $\sigma_2$ representing obstacles $\Omega_1$ and $\Omega_2$, we would typically find the SDF of $\Omega_1 \cup \Omega_2$ by taking their minimum $\sigma(p) = \min(\sigma_1(p), \sigma_2(p))$. However, this poses problems of differentiability, which is necessary for IPOPT to solve our optimization problem. We resolve this issue by computing a twice-differentiable lower-bound approximation of the minimum function as shown in \cite{quilez_sdf}, which is necessary because the ZBF constraint contains a first-order derivative of ZBF $h(x)$.

\begin{align*}
    \smin(d_1, d_2, k) = &\min(d_1, d_2) \\
    &-k\left(\frac{1}{6}\right)\left(\frac{\max(k - |d_1 - d_2|, 0.0)}{k}\right)^3
\end{align*}

\noindent where $k$ can be interpreted as a blending radius. With repeated application of the $\smin$ operation, we can have a twice-differentiable SDF union of all the obstacles known in the workspace.

We then state the following lemma, given the informal arguments above about our current ZBF constructions:

\begin{lem}
Let $\dot x = g(x)$ be a DMP, $\gamma > 0$, $h: \mathbb R^3 \to \mathbb R$ be a ZBF, and $\mathcal C = \{x \in \mathbb R^3 : h(x) > 0\}$. If there exists $\zeta_1, ..., \zeta_n$ such that the constrained DMP $\dot x = \tilde g(x)$ admits the ZBF inequality  $\dot h(x) + \gamma h(x) \geq 0$ along trajectory $x(t)$, then  CDMP $\tilde g$'s trajectory x(t) is forward-invariant in $\mathcal C$.

\begin{proof}
Application of Ames et al.'s Proposition B.1 \cite{ames2016control}.
\end{proof}
\end{lem}

\noindent As an implication of the above lemma, we can claim that if our solver is able to find feasible solution to~\eqref{eqn:cdmp}-~\eqref{eqn:init_condition} to satisfy the ZBF condition, the optimized CDMP trajectory is guaranteed to stay within the safety set.



\section{Experiments and Results}\label{sec:results}
We tested our algorithm in numerical and physical experiments to understand its efficacy. The numerical examples are used to verify that our method could be used with a wide variety of ZBFs (and thus safety sets), test our method's performance against existing methods for set-invariance of DMPs, and understand its computational efficiency. Our physical experiments showed feasibility of our algorithm in a real-world setting. All the experiments presented in the next Section are performed using a Python-based interface 
to IPOPT presented in ~\cite{DBLP:journals/corr/abs-2106-03220}.

\subsection{Numerical Experiments}\label{subsec:numerical_exps}

\begin{figure}
    \centering
    \includegraphics[width=0.40\textwidth]{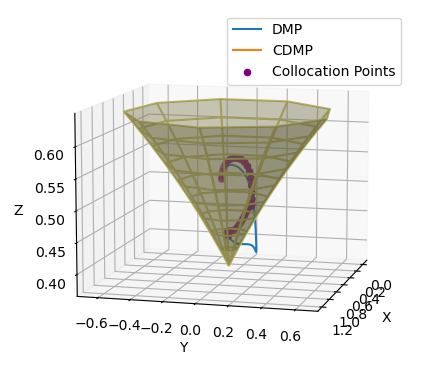}
    \caption{CDMP trajectory constrained to lie within a cone. The ZBF was constructed using the SDFs of a circles arranged along an axis, their radii paramaterized by $z$. (Best seen in color.)}
    \label{fig:cone_stay}
\end{figure}

We first present numerical experiments to verify that CDMPs are  forward-invariant in the safety set as defined by the ZBF. We recorded an initial demonstration end-effector trajectory kinesthetically taught by a human expert in $\mathbb R^3$ (see Figure~\ref{fig:cone_stay}). 
Then, we impose a runtime constraint that the robot trajectory needs to follow state constraints of the type $x(t)\in (\mathcal{X}(t))^c$, where $\mathcal{X}(t)$ could be non-compact and could be defined by a user depending on a task of interest. This kind of constraint satisfaction arises in robotic tasks where the task-specific constraint requires that the robot should always stay in a certain set during the task. An example of such a task could be an assembly task where the end-effector needs to always stay in a pre-defined set with respect to the part positions. As could be seen in Figure~\ref{fig:cone_stay}, the CDMP trajectory is able to satisfy the specified constraint. Furthermore, we can also easily specify and satisfy time-dependent state bounds for the trajectory which might be useful for a lot of tasks. However, we skip these results for brevity.


\subsection{Comparison to Dynamic Artificial Potential Fields}\label{sec:apf}
\begin{figure}
    \centering
    \includegraphics[width=0.40\textwidth]{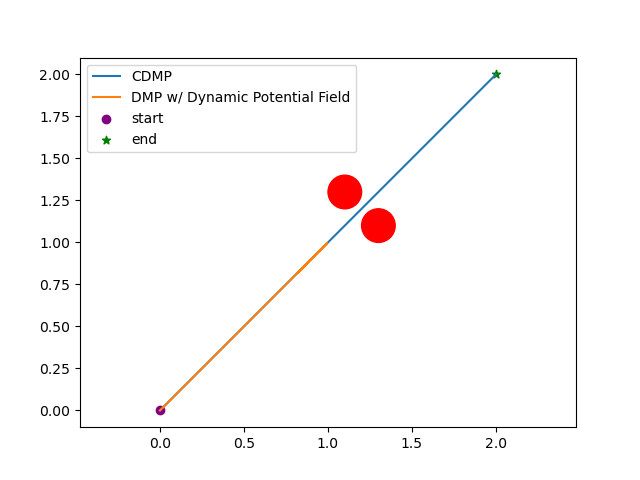}
    \caption{CDMP and DMP with a dynamic potential field learned from generated straight-line demonstration. The former is able to reach the goal, while the latter is caught in a local minimum. (Best seen in color.)}
    \label{fig:cdmp_vs_apf}
\end{figure}

We compare our method to the dynamic (velocity-dependant) artificial potential field method introduced in \cite{park2008movement} using a generated straight-line trajectory. As with many potential field methods, it is not difficult to construct an adversarial obstacle configuration that causes the DMP to be trapped in a local minimum (see Fig. \ref{fig:cdmp_vs_apf}). Even with these obstacles, our CDMP method is able to successfully reach the goal. This demonstrates that our method is fundamentally different from DMPs perturbed with potential field methods, and can allow for workspace constraint satisfaction in a more principled manner.

\subsection{Computational Efficiency}\label{sec:compute_efficiency}

\begin{figure}
    \centering
    \includegraphics[width=0.40\textwidth]{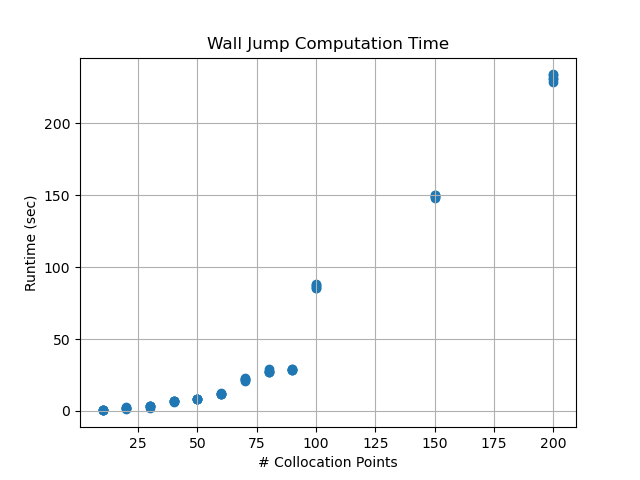}
    \caption{Experimental runtime of wall-jump problem.}
    \label{fig:comp_exp}
\end{figure}

\begin{figure}
    \centering
    \includegraphics[width=0.49\textwidth]{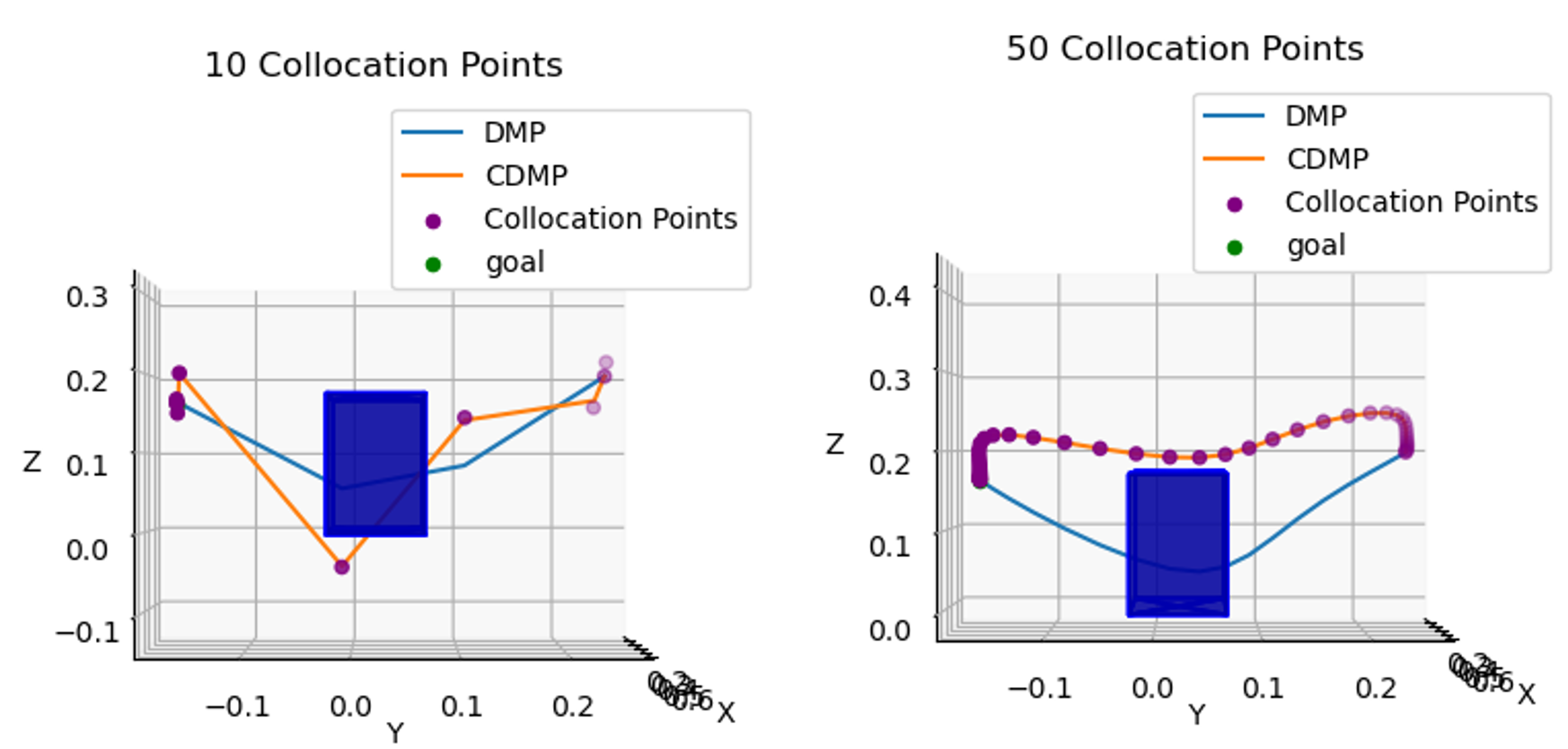}
    \caption{While $n=10$ collocation points was much faster to compute (Fig. \ref{fig:comp_exp}), the trajectory was still in collision with the obstacle since the ZBF was too coarsely enforced relative to $n=50$.}
    \label{fig:collision_fidelity}
\end{figure}

%
Apart from the complexity of the constraints, the computational time for CDMP depends on the size of the NLP created using the original demonstration trajectory. The size of the problem depends on the number of collocation points for the trajectory.  We note that constraints are only enforced at the collocation points when the CDMP optimization problem is transformed into a finite dimensional NLP, so the constraints may be violated between the collocation points if the discretization is too coarse. We evaluate the computational efficiency of our algorithm by varying the number of collocation points during optimization (Fig. \ref{fig:comp_exp}). All experiments were run using unoptimized code on a Lenovo Thinkpad equipped with an Intel i7-8565U processor and 16Gb of RAM. While fewer collocation points result in faster compute times, the trajectory in between collocation points is more likely to intersect with obstacles (Fig. \ref{fig:collision_fidelity}). As a result of this trade-off, the correct number of collocation points should be chosen based on the problem. Figure~\ref{fig:comp_exp} shows a non-linear trend between the number of collocation points and the computational time. This is not surprising since we usie an interior-point method (IPM) for a non-convex optimization, where the computational time can increase sharply with problem size.


\begin{figure*}
    \centering
    \includegraphics[width=0.75\textwidth]{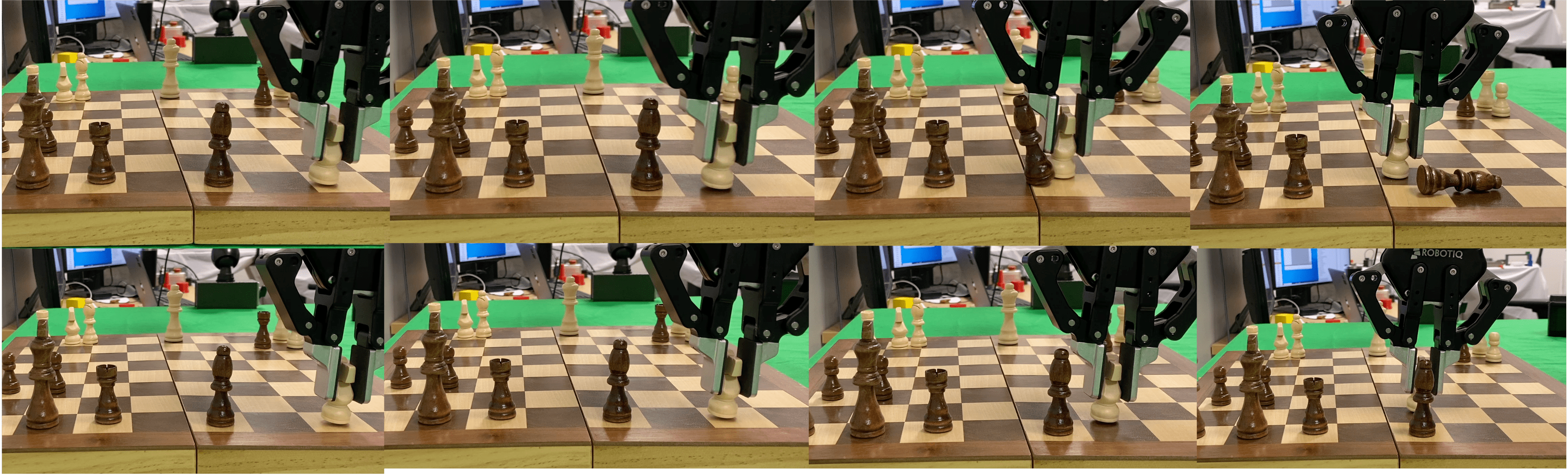}
    \caption{This shows movement of pieces on a chess board using CDMP. The top row shows that the original DMP collides with an obstacle on the way while the CDMP (bottom row) is able to avoid the obstacle during movement.}
    \label{fig:chessboard_cdmp}
\end{figure*}

\subsection{Physical Robot Experiments}
\begin{figure}
        \centering
        \begin{subfigure}[b]{0.45\textwidth}
            \centering
            \includegraphics[width=\textwidth]{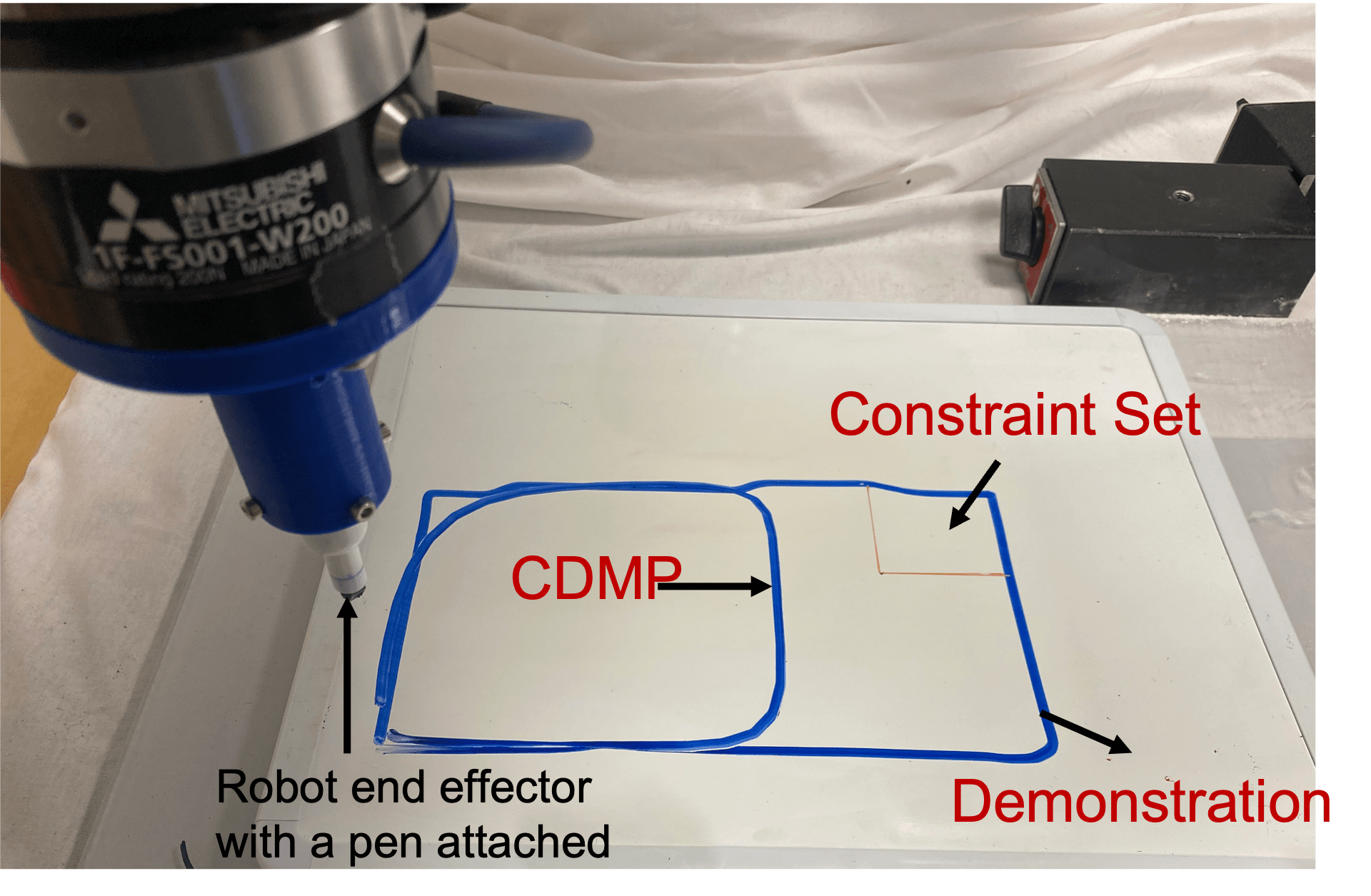}
            \caption{Drawing a rectangle.}
            \label{fig:draw_rect}
        \end{subfigure}
        \hfill
        \begin{subfigure}[b]{0.45\textwidth}  
            \centering 
            \includegraphics[width=\textwidth]{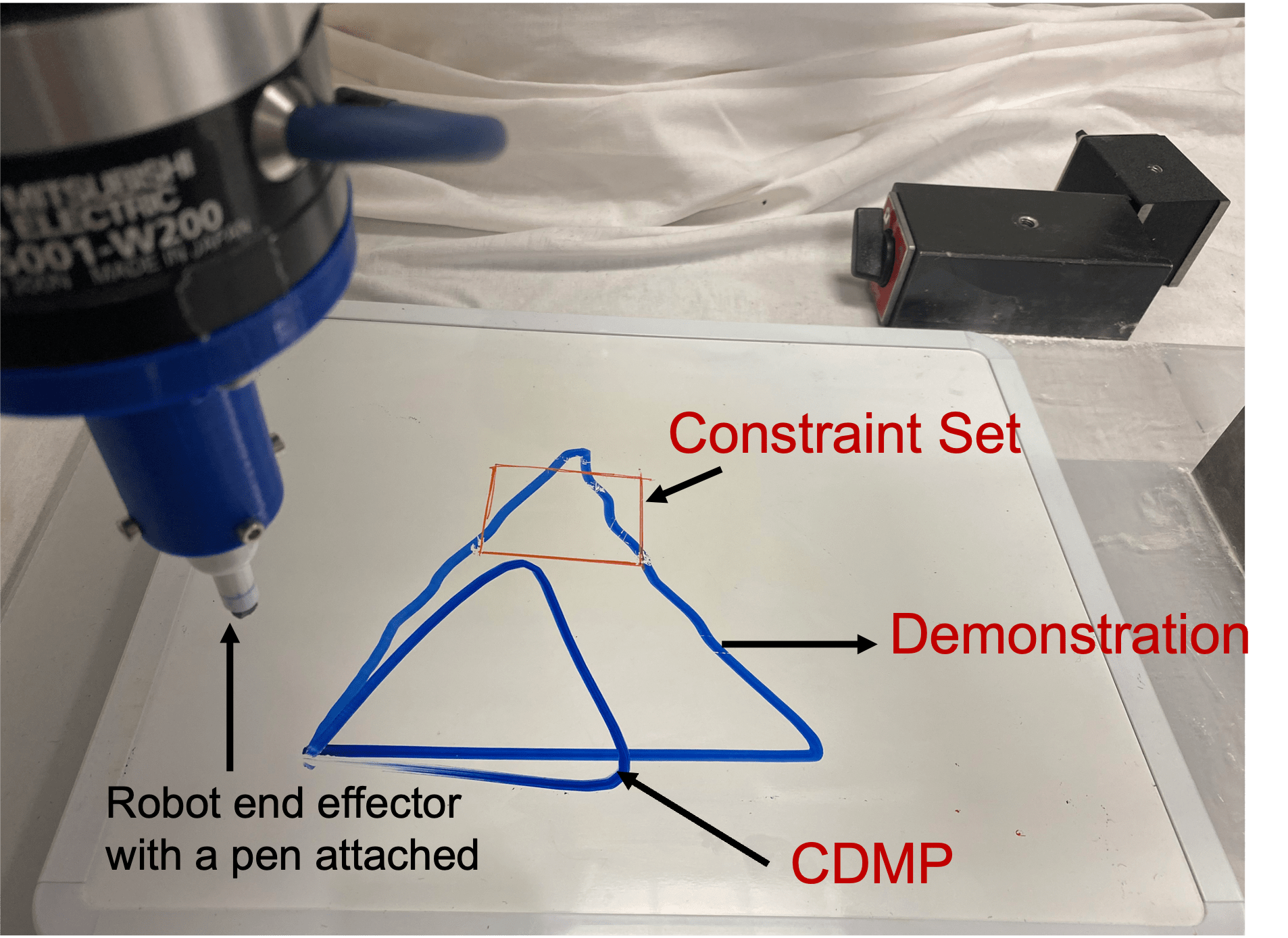}
            \caption{Drawing a triangle.}
            \label{fig:draw_triangle}
        \end{subfigure}
        \caption{The proposed CDMP algorithm is successfully able to draw demonstrated shapes in the presence of constraints.}
        \label{fig:drawing_exp}
\end{figure}

We tested our CDMP on two different physical robots in three different domains: a wall-hopping domain (see Figure~\ref{fig:cdmp_obstacle_avoidance}), a chess-piece moving domain (see Figure~\ref{fig:chessboard_cdmp}), and a whiteboard drawing domain (see Figure~\ref{fig:drawing_exp}). We test the overall efficacy of our approach in the wall-hopping and chess-moving domains, and the drawing domain tests the CDMP's ability to preserve the qualitative geometry of the original DMP trajectory. A video showing the implementation of the proposed algorithm in these environments could be found here \url{https://youtu.be/hJegJJkJfys}.

In the wall and chess domains, we use a Kinova Gen3 robot. In these experiments, a user kinesthetically demonstrates the robot a movement in a workspace free of obstacles. To test the CDMP's ability to generate collision-free trajectories, we introduce new obstacles in the environment. The environment for the wall-hopping domain can be seen in Figure~\ref{fig:cdmp_obstacle_avoidance}, where the CDMP finds a collision-free trajectory. Similarly, for the chessboard environment in Figure~\ref{fig:chessboard_cdmp}, we provided an initial demonstration to move the knight piece. While DMP (top row) fails in the presence of a new obstacle, CDMP (bottom row) is able to compute a new non-colliding trajectory. More experiments are shown in the supplementary video.

For the drawing experiments, we use a Mitsubishi Electric
Assista industrial manipulator arm equipped with a force/torque (F/T) sensor at the wrist. The F/T sensor is used to design an admittance controller~\cite{jha2022design} to move the robot in a kinesthetic teaching mode for contact-rich tasks. This controller is used to provide demonstrations on a white-board to draw several shapes such as a rectangle and a triangle (see Figure~\ref{fig:drawing_exp}). Then, we add some constraint sets in the original shapes, which the robot should avoid while trying to re-create the shape. The initial demonstration and the final CDMP trajectories for the experiments could be seen in Figure~\ref{fig:drawing_exp}. In all cases, the CDMP is able to find feasible solutions while preserving the geometric shape of the drawings in both the experiments.

\section{Conclusion and Future Work}\label{sec:conclusion}
In this paper, we propose a method for incorporating operational constraints while generating trajectories for generalizing to novel initial and goal locations. Our algorithm, constrained dynamic movement primitives, can incorporate task and environmental constraints when generalizing to novel conditions. The proposed CDMP was demonstrated on several examples for collision avoidance in the presence of obstacles of different shapes and sizes, and also preserves the qualitative geometry of the trajectory. 



Even though the formulation can, in theory, include complex operational constraints like self-collision, joint limits, etc., we need computationally efficient and smooth barrier function representation of these constraints to be able to compute safe trajectories. In the future, we would like to use the proposed formulation to consider more constraints past end-effector safety sets that can be more easily expressed in different parameterizations of the robot configuration space, such as joint limits, self-collision avoidance, etc. These constraints result in a complex, constrained optimization problem which needs additional, non-trivial work to find a suitable zeroing barrier function. Another limitation is the computational efficiency with respect to the size of the problem. In this paper, we use IPOPT to solve the resulting optimization which can not make efficient use of warm starting. In our future work, we plan to integrate our problem with an active-set solver to improve the computational efficiency.


\bibliographystyle{IEEEtran}
\bibliography{references}

\begin{thebibliography}{10}
\providecommand{\url}[1]{#1}
\csname url@samestyle\endcsname
\providecommand{\newblock}{\relax}
\providecommand{\bibinfo}[2]{#2}
\providecommand{\BIBentrySTDinterwordspacing}{\spaceskip=0pt\relax}
\providecommand{\BIBentryALTinterwordstretchfactor}{4}
\providecommand{\BIBentryALTinterwordspacing}{\spaceskip=\fontdimen2\font plus
\BIBentryALTinterwordstretchfactor\fontdimen3\font minus
  \fontdimen4\font\relax}
\providecommand{\BIBforeignlanguage}[2]{{%
\expandafter\ifx\csname l@#1\endcsname\relax
\typeout{** WARNING: IEEEtran.bst: No hyphenation pattern has been}%
\typeout{** loaded for the language `#1'. Using the pattern for}%
\typeout{** the default language instead.}%
\else
\language=\csname l@#1\endcsname
\fi
#2}}
\providecommand{\BIBdecl}{\relax}
\BIBdecl

\bibitem{argall2009survey}
B.~D. Argall, S.~Chernova, M.~Veloso, and B.~Browning, ``A survey of robot
  learning from demonstration,'' \emph{Robotics and autonomous systems},
  vol.~57, no.~5, pp. 469--483, 2009.

\bibitem{ravichandar2020recent}
H.~Ravichandar, A.~S. Polydoros, S.~Chernova, and A.~Billard, ``Recent advances
  in robot learning from demonstration,'' \emph{Annual Review of Control,
  Robotics, and Autonomous Systems}, vol.~3, pp. 297--330, 2020.

\bibitem{schaal2006dynamic}
S.~Schaal, ``Dynamic movement primitives-a framework for motor control in
  humans and humanoid robotics,'' in \emph{Adaptive motion of animals and
  machines}.\hskip 1em plus 0.5em minus 0.4em\relax Springer, 2006, pp.
  261--280.

\bibitem{saveriano2021dynamic}
M.~Saveriano, F.~J. Abu-Dakka, A.~Kramberger, and L.~Peternel, ``Dynamic
  movement primitives in robotics: A tutorial survey,'' \emph{arXiv preprint
  arXiv:2102.03861}, 2021.

\bibitem{5686298}
K.~Muelling, J.~Kober, and J.~Peters, ``Learning table tennis with a mixture of
  motor primitives,'' in \emph{2010 10th IEEE-RAS International Conference on
  Humanoid Robots}, 2010, pp. 411--416.

\bibitem{mulling2013learning}
K.~M{\"u}lling, J.~Kober, O.~Kroemer, and J.~Peters, ``Learning to select and
  generalize striking movements in robot table tennis,'' \emph{The
  International Journal of Robotics Research}, vol.~32, no.~3, pp. 263--279,
  2013.

\bibitem{9838102}
D.~K. Jha, D.~Romeres, W.~Yerazunis, and D.~Nikovski, ``Imitation and
  supervised learning of compliance for robotic assembly,'' in \emph{2022
  European Control Conference (ECC)}, 2022, pp. 1882--1889.

\bibitem{sharma2019learning}
M.~Sharma, K.~Zhang, and O.~Kroemer, ``Learning semantic embedding spaces for
  slicing vegetables,'' \emph{arXiv preprint arXiv:1904.00303}, 2019.

\bibitem{ames2016control}
A.~D. Ames, X.~Xu, J.~W. Grizzle, and P.~Tabuada, ``Control barrier function
  based quadratic programs for safety critical systems,'' \emph{IEEE
  Transactions on Automatic Control}, vol.~62, no.~8, pp. 3861--3876, 2016.

\bibitem{wachter2006implementation}
A.~W{\"a}chter and L.~T. Biegler, ``On the implementation of an interior-point
  filter line-search algorithm for large-scale nonlinear programming,''
  \emph{Mathematical programming}, vol. 106, no.~1, pp. 25--57, 2006.

\bibitem{DBLP:journals/corr/abs-2106-03220}
A.~U. Raghunathan, D.~K. Jha, and D.~Romeres, ``{PYROBOCOP} : Python-based
  robotic control {\&} optimization package for manipulation and collision
  avoidance,'' \emph{CoRR}, vol. abs/2106.03220, 2021.

\bibitem{rai2014learning}
A.~Rai, F.~Meier, A.~Ijspeert, and S.~Schaal, ``Learning coupling terms for
  obstacle avoidance,'' in \emph{2014 IEEE-RAS International Conference on
  Humanoid Robots}.\hskip 1em plus 0.5em minus 0.4em\relax IEEE, 2014, pp.
  512--518.

\bibitem{park2008movement}
D.-H. Park, H.~Hoffmann, P.~Pastor, and S.~Schaal, ``Movement reproduction and
  obstacle avoidance with dynamic movement primitives and potential fields,''
  in \emph{Humanoids 2008-8th IEEE-RAS International Conference on Humanoid
  Robots}.\hskip 1em plus 0.5em minus 0.4em\relax IEEE, 2008, pp. 91--98.

\bibitem{tan2011potential}
H.~Tan, E.~Erdemir, K.~Kawamura, and Q.~Du, ``A potential field method-based
  extension of the dynamic movement primitive algorithm for imitation learning
  with obstacle avoidance,'' in \emph{2011 IEEE International Conference on
  Mechatronics and Automation}.\hskip 1em plus 0.5em minus 0.4em\relax IEEE,
  2011, pp. 525--530.

\bibitem{stulp2009compact}
F.~Stulp, E.~Oztop, P.~Pastor, M.~Beetz, and S.~Schaal, ``Compact models of
  motor primitive variations for predictable reaching and obstacle avoidance,''
  in \emph{2009 9th IEEE-RAS International Conference on Humanoid
  Robots}.\hskip 1em plus 0.5em minus 0.4em\relax IEEE, 2009, pp. 589--595.

\bibitem{sobti2021sampling}
S.~Sobti, R.~Shome, S.~Chaudhuri, and L.~E. Kavraki, ``A sampling-based motion
  planning framework for complex motor actions,'' in \emph{2021 IEEE/RSJ
  International Conference on Intelligent Robots and Systems (IROS)}.\hskip 1em
  plus 0.5em minus 0.4em\relax IEEE, 2021, pp. 6928--6934.

\bibitem{ames2014rapidly}
A.~D. Ames, K.~Galloway, K.~Sreenath, and J.~W. Grizzle, ``Rapidly
  exponentially stabilizing control lyapunov functions and hybrid zero
  dynamics,'' \emph{IEEE Transactions on Automatic Control}, vol.~59, no.~4,
  pp. 876--891, 2014.

\bibitem{ames2019control}
A.~D. Ames, S.~Coogan, M.~Egerstedt, G.~Notomista, K.~Sreenath, and P.~Tabuada,
  ``Control barrier functions: Theory and applications,'' in \emph{2019 18th
  European control conference (ECC)}.\hskip 1em plus 0.5em minus 0.4em\relax
  IEEE, 2019, pp. 3420--3431.

\bibitem{romdlony2014uniting}
M.~Z. Romdlony and B.~Jayawardhana, ``Uniting control lyapunov and control
  barrier functions,'' in \emph{53rd IEEE Conference on Decision and
  Control}.\hskip 1em plus 0.5em minus 0.4em\relax IEEE, 2014, pp. 2293--2298.

\bibitem{tedrake2010lqr}
R.~Tedrake, I.~R. Manchester, M.~Tobenkin, and J.~W. Roberts, ``Lqr-trees:
  Feedback motion planning via sums-of-squares verification,'' \emph{The
  International Journal of Robotics Research}, vol.~29, no.~8, pp. 1038--1052,
  2010.

\bibitem{khansari2011learning}
S.~M. Khansari-Zadeh and A.~Billard, ``Learning stable nonlinear dynamical
  systems with gaussian mixture models,'' \emph{IEEE Transactions on Robotics},
  vol.~27, no.~5, pp. 943--957, 2011.

\bibitem{khansari2014learning}
------, ``Learning control lyapunov function to ensure stability of dynamical
  system-based robot reaching motions,'' \emph{Robotics and Autonomous
  Systems}, vol.~62, no.~6, pp. 752--765, 2014.

\bibitem{figeuroa2017physically}
\BIBentryALTinterwordspacing
N.~Figueroa and A.~Billard, ``A physically-consistent bayesian non-parametric
  mixture model for dynamical system learning,'' in \emph{Proceedings of The
  2nd Conference on Robot Learning}, ser. Proceedings of Machine Learning
  Research, A.~Billard, A.~Dragan, J.~Peters, and J.~Morimoto, Eds.,
  vol.~87.\hskip 1em plus 0.5em minus 0.4em\relax PMLR, 29--31 Oct 2018, pp.
  927--946. [Online]. Available:
  \url{https://proceedings.mlr.press/v87/figueroa18a.html}
\BIBentrySTDinterwordspacing

\bibitem{ravichandar2017learning}
H.~Ravichandar, I.~Salehi, and A.~Dani, ``Learning partially contracting
  dynamical systems from demonstrations,'' in \emph{Conference on Robot
  Learning}.\hskip 1em plus 0.5em minus 0.4em\relax PMLR, 2017, pp. 369--378.

\bibitem{notomista2021safety}
G.~Notomista and M.~Saveriano, ``Safety of dynamical systems with multiple
  non-convex unsafe sets using control barrier functions,'' \emph{IEEE Control
  Systems Letters}, vol.~6, pp. 1136--1141, 2021.

\bibitem{khansari2012dynamical}
S.~M. Khansari-Zadeh and A.~Billard, ``A dynamical system approach to realtime
  obstacle avoidance,'' \emph{Autonomous Robots}, vol.~32, no.~4, pp. 433--454,
  2012.

\bibitem{ohnishi2021constraint}
M.~Ohnishi, G.~Notomista, M.~Sugiyama, and M.~Egerstedt, ``Constraint learning
  for control tasks with limited duration barrier functions,''
  \emph{Automatica}, vol. 127, p. 109504, 2021.

\bibitem{jones20063d}
M.~W. Jones, J.~A. Baerentzen, and M.~Sramek, ``3d distance fields: A survey of
  techniques and applications,'' \emph{IEEE Transactions on visualization and
  Computer Graphics}, vol.~12, no.~4, pp. 581--599, 2006.

\bibitem{quilez_sdf}
\BIBentryALTinterwordspacing
I.~Quilez, ``3d distance functions.'' [Online]. Available:
  \url{https://iquilezles.org/www/articles/distfunctions/distfunctions.htm}
\BIBentrySTDinterwordspacing

\bibitem{jha2022design}
D.~K. Jha, D.~Romeres, S.~Jain, W.~Yerazunis, and D.~Nikovski, ``Design of
  adaptive compliance controllers for safe robotic assembly,'' \emph{arXiv
  preprint arXiv:2204.10447}, 2022.

\end{thebibliography}
\end{document}